%% file: main.tex
\documentclass[11pt]{article}

\input{macros.tex}

\title{
    Privacy-Preserving Race/Ethnicity Estimation for Algorithmic Bias Measurement in the U.S.
}
\author{
    Saikrishna Badrinarayanan\thanks{These two authors contributed equally to this work.} \and  Osonde Osoba\footnotemark[1] \and Miao Cheng \and Ryan Rogers \and Sakshi Jain \and Rahul Tandra \and Natesh S. Pillai 
}
\date{ LinkedIn}

\begin{document}
\maketitle

\begin{abstract}
    
    AI fairness measurements, including tests for equal treatment, often take the form of disaggregated evaluations of AI systems.
    Such measurements are an important part of Responsible AI operations.
    These measurements compare system performance across demographic groups or sub-populations and typically require member-level demographic signals such as gender, race, ethnicity, and location. 
    However, sensitive member-level demographic attributes like race and ethnicity can be challenging to obtain and use due to platform choices, legal constraints, and cultural norms.
    In this paper, we focus on the task of enabling AI fairness measurements on race/ethnicity for \emph{U.S. LinkedIn members} in a privacy-preserving manner.
    We present the Privacy-Preserving Probabilistic Race/Ethnicity Estimation (PPRE) method for performing this task. 
    PPRE combines the Bayesian Improved Surname Geocoding (BISG) model, a sparse LinkedIn survey sample of self-reported demographics, and privacy-enhancing technologies like secure two-party computation and differential privacy to enable meaningful fairness measurements while preserving member privacy.
    We provide details of the PPRE method and its privacy guarantees. 
    We then illustrate sample measurement operations. We conclude with a review of open research and engineering challenges for expanding our privacy-preserving fairness measurement capabilities. 
\end{abstract}


\input{introduction.tex}

\input{building-blocks-f-audits.tex}

\input{building-blocks-privacy.tex}

\input{ppre-details.tex}

\input{implementation.tex}

\input{sample-analyses.tex}

\input{conclusion.tex}

\printbibliography

\end{document}

%% file: macros.tex
\def\ShowAuthNotes{1}

\ifnum\ShowAuthNotes=1
	\newcommand{\sai}[1]{\textcolor{blue}{[{\bf Sai:} {#1}]}}
	\newcommand{\osonde}[1]{\textcolor{blue}{[{\bf Osonde:} {#1}]}}
        \newcommand{\rynote}[1]{\textcolor{red}{[{\bf Ryan:} {#1}]}}
        \newcommand{\miao}[1]{\textcolor{blue}{[{\bf Miao:} {#1}]}}
\else
	\newcommand{\sai}[1]{}
	\newcommand{\osonde}[1]{}
        \newcommand{\rynote}[1]{}
        \newcommand{\miao}[1]{}

\fi

\usepackage{fullpage}
\usepackage{amsthm}
\usepackage{amsmath}
\usepackage{amsfonts}
\usepackage{amssymb}

\usepackage{hyperref}
\hypersetup{
	colorlinks=true,
	urlcolor=blue   
}

\usepackage{graphicx, url, xspace}
\usepackage[justification=centering]{caption}

\usepackage[backend=biber, sorting=none]{biblatex}
\newcommand{\sk}{\mathsf{sk}}
\newcommand{\pk}{\mathsf{pk}}
\newcommand{\ct}{\mathsf{ct}}
\newcommand{\M}{\mathcal{M}}
\newcommand{\G}{\mathcal{G}}
\newcommand{\secr}{\lambda}
\newcommand{\RO}{\mathsf{RO}}
\newcommand{\ahesk}{\mathsf{sk_{HE}}}
\newcommand{\ahepk}{\mathsf{pk_{HE}}}
\newcommand{\symsk}{\mathsf{sk_{Sym}}}
\newcommand{\rand}{\mathsf{rand}}
\newcommand{\cS}{\mathcal{S}}
\newcommand{\bT}{\mathsf{T}}

\newcommand{\ahe}{\mathsf{HE}}
\newcommand{\ahekeygen}{\mathsf{HE.KeyGen}}
\newcommand{\aheenc}{\mathsf{HE.Enc}}
\newcommand{\aheadd}{\mathsf{HE.Add}}
\newcommand{\ahedec}{\mathsf{HE.Dec}}

\newcommand{\com}{\mathsf{Com}}
\newcommand{\comkeygen}{\mathsf{Com.KeyGen}}
\newcommand{\comenc}{\mathsf{Com.Enc}}
\newcommand{\comdec}{\mathsf{Com.Dec}}

\newcommand{\symkeygen}{\mathsf{Sym.KeyGen}}
\newcommand{\symenc}{\mathsf{Sym.Enc}}
\newcommand{\symdec}{\mathsf{Sym.Dec}}

\newcommand{\namedref}[2]{\hyperref[#2]{#1~\ref*{#2}}\xspace}
\newcommand{\sectionref}[1]{\namedref{Section}{sec:#1}}
\newcommand{\figureref}[1]{\namedref{Figure}{fig:#1}}
\newcommand{\memid}{\mathsf{mem_{id}}}
\newcommand{\out}{\mathsf{output}}
\newcommand{\racep}{\vec{\mathsf{race_p}}}
\newcommand{\values}{\vec{\mathsf{values}}}

%% file: introduction.tex
\section{Introduction}

LinkedIn leverages AI at scale to help its community of members use the platform more effectively to gain access to economic opportunity by connecting with peers, getting access to matched job recommendations, finding more relevant content in their feed, etc.
In 2023, LinkedIn took the step of sharing its Responsible AI Principles~\cite{LawitXu_2023} which are intended to guide its use of AI. 
These principles include a commitment to work to ensure we treat members equitably and we do not amplify pre-existing societal disparities.
The principles include a commitment to ``Promote Fairness and Inclusion: We work to ensure that our use of AI benefits all members fairly, without causing or amplifying unfair bias.''
Measurement is the first step for achieving this commitment to identify and address wrongful biases present in our AI systems. 
Performing these measurements will depend on our ability to first measure and understand the impact of our AI systems along key demographic dimensions, including \emph{race and ethnicity}. 

We aim to focus on addressing wrongful biases or disparities that are within an AI-deploying institution's sphere of control, i.e. wrongful biases in the institution's AI systems.
AI fairness measurements are the first step for addressing any wrongful biases present in AI systems.
This includes measurements for race and ethnicity in the U.S.
Such AI fairness measurements for race and ethnicity are thus an essential part of  Responsible AI (RAI) operational programs within the U.S.
However, the absence of data on race/ethnicity has the effect of making race \& ethnicity fairness measurements difficult to execute. 

A typical fairness measurement includes disaggregated evaluations~\cite{barocas2021} of the model or system under investigation.
This consists of quantifying and comparing the performance of the evaluation target using input samples from different demographic categories.
These types of group-wise statistical comparisons are the backbone of fairness or equity analyses in RAI programs.
Generating these disaggregation evaluation metrics typically requires the use of \emph{member-level} demographic data.
Such demographic data on race/ethnicity can be challenging to obtain and use due to platform choices, regulatory constraints, and cultural norms.
For example, members may also have misgivings about public disclosures of their race/ethnicity information. 
And in any case, both personal data best practices and privacy regulations promote the practice of data minimization\footnote{Data minimization mandates using the minimum data required to accomplish a given purpose.} around potentially sensitive personal data such as race/ethnicity. 

For example, in the U.S., race and ethnicity attributes are only available for about $6\%$ of the U.S. LinkedIn member base via self-reported demographic information on the LinkedIn Self-ID survey\footnote{\url{https://members.linkedin.com/equal-access}}. 
This means we do not have enough data to do AI fairness measurements based on members' race/ethnicity groups. 
Self-ID survey samples can be used to enable limited AI measurements. 
measurement results based solely on Self-ID survey samples, however, do not generalize to broader LinkedIn member sub-populations because of the effects of Self-ID's non-random sampling.
There is some selection bias in the Self-ID survey as different types of LinkedIn members choose whether or not to respond to the survey for different reasons. 
To extend the coverage of our equity measurements beyond the Self-ID sample, we considered new approaches in the U.S. 
including:
\begin{itemize}
    \item Constructing and relying on ``direct association'' tables (DA). This involves the use of members' expressed affiliations to estimate demographic groupings.
    \item `Debiasing` the Self-ID survey. This involves the application of standard causal inference techniques based on propensity scores to attempt to remove selection bias effects from Self-ID-derived statistical estimates.
\end{itemize}
These explorations were not pursued by us because neither of these paths were viable.
The first method did not meet our high standard for member transparency and control over their personal data because we did not want to infer and then ``assign'' a race or ethnicity to any member in our systems. 
Furthermore, group affiliations are not necessarily reliable signals of belonging to a particular demographic background. 
In the second case, we did not believe the ``debiased'' estimates of fairness measurements could be made reliable and robust enough to drive decisions.

This leads us to the central problem we address in this paper: 
\emph{monitoring our AI systems for fairness according to U.S. members' race and ethnicity requires knowledge of those members’ race and ethnicity; but platform data on race and ethnicity attributes is severely limited.} 

\paragraph{Principles}
We decided to focus this work on the U.S. Our design principles can be summarized as follows:
\begin{itemize}
    \item \textbf{Impact:} Race/ethnicity demographic information must be comprehensive and useful enough to enable equal treatment measurements with respect to race and ethnicity at the aggregate level.
\item \textbf{Privacy:} The measurement test must have privacy by design at its core, including:
\begin{itemize}
    \item Data minimization: We must use (or “process”) the least amount of personal data needed to achieve our objective and respect the sensitivities members may have around race/ethnicity information, including avoiding LinkedIn “assigning,” saving, or disclosing race/ethnicity information. 
\item No individual race/ethnicity assignment: We will avoid individual assignment of a single race or ethnicity category to members.
\item Strong Security: We will implement privacy and security protections to prevent unauthorized access (internal or external) to this measurement workflow (e.g. efforts to prevent reassociation of an estimate calculation to an identifiable person)
\item Transparency and Member Control: We must be transparent about the personal data we are using for race/ethnicity measurement tests and provide appropriate controls to members. So, even though a single race/ethnicity will not be assigned to any member, members can opt-out of having their personal data used for the purpose of this race/ethnicity measurement test.
\end{itemize}
\end{itemize}

\subsection{Our Results}
Adhering to these principles will enable us to make meaningful progress towards our commitment to monitor LinkedIn’s AI systems for unintended bias without compromising member privacy or control over their race/ethnicity identity. To address this challenge, we designed the Privacy-Preserving Probabilistic Race/Ethnicity Estimation (PPRE) method for conducting privatized race \& ethnicity fairness measurements on U.S. member bases. Our system combines the below techniques from responsible AI and privacy in a novel manner:
\begin{itemize}
    \item the Bayesian Improved Surname Geocoding (BISG) model~\cite{elliott2009,imai2022}, a U.S.-Census-normed Bayesian model for estimating race/ethnicity probabilities;
    \item the sparse LinkedIn Self-ID survey of self-reported demographics; and 
    \item Privacy enhancing technologies, including secure two-party computation (2PC) and differential privacy (DP)
\end{itemize}

\paragraph{Design Evolution Process.} 
We arrived at our solution after a few rounds of iterations with various stakeholders, including a discussion with experts in this space from academia and industry at a LinkedIn-hosted workshop~\cite{hai-wrkshp2023}.
Our initial approach was to consider writing all data to a common location on disk and then performing the measurements in this isolated environment. 
We quickly dropped that since it would require persisting and computing directly on sensitive data violating the constraints we laid out earlier. 
The next iteration was to consider privatizing Self-ID data using local differential privacy, store them encrypted at rest, and then, via strong access controls, decrypt and compute measurements on sensitive plaintext data. 
Once again, this did not satisfy our requirements/privacy guarantees to prevent new disclosures of members' sensitive data and that there should be no identifiable plaintext analysis.
This iteration was unsatisfactory for another reason; the solution here would feature a persistently stored table in which some member data would be deliberately falsified due to the application of local DP.
We aim to avoid PPRE solutions or applications that lead to a potentially persistent association of members with incorrect or false labels. 

We then explored the idea of performing a secure two-party computation protocol to ensure that neither the tester nor the test client learns sensitive information. For each measurement, we would generate ephemeral race/ethnicity data on the fly, immediately encrypt it and decrypt only aggregate data at the end of the test. However, keeping performance and scale in mind, we did not consider a general purpose protocol -- for instance, applying fully homomorphic encryption to the entire dataset. To offer a reasonable trade-off between utility, computation cost, communication cost, and privacy, we had to carefully incorporate and intertwine the right technical tools to build our custom secure computation protocol in combination with privatizing the dataset using differential privacy. We elaborate more on our design in the next sections.

Our final system achieves the following:
\begin{enumerate}
    \item \textbf{No one is assigned a race/ethnicity:} We calculate only probabilistic estimates that are ephemeral on the member level; during each measurement, they are computed on demand, never stored on disk, encrypted by design and aggregated. No member level estimations are used to alter how our AI systems specifically treat that member.

\item \textbf{Secure exchange of data:} Different system providers (the tester and the test client; details elaborated later) of relevant data within LinkedIn exchange only encrypted data that is not decryptable by the other or to an eavesdropper.

\item \textbf{No new disclosures of a member’s sensitive personal data:} The secure two-party computation protocol ensures that at any point during the execution of this measurement or after, neither the measurement executor nor the measurement client gains knowledge of the tuple (member ID, race/ethnicity probabilities, test features). 

\item \textbf{Supporting deniability on targeted race/ethnicity identification:} Differential privacy helps prevent internal teams that perform algorithmic bias testing from being able to re-identify members’ race/ethnicity estimations or Self-ID values by looking at the output from multiple test flows. In addition, we mandate a governance approval process before embarking on any new measurements. 

\item \textbf{Retain only aggregate data:} We save only aggregate privatized information at the end of the test and delete all member level (even encrypted) data as soon as it has been processed. We only use anonymized aggregated data to test our AI systems for equal treatment by race/ethnicity.
\item \textbf{Member Control:} We enable members to opt-out of having their personal data used for measurement through their data privacy settings.
\end{enumerate}

\subsection{Related Prior Art}
The PPRE method is a hybridization of methods from multiple disciplines.
The component methods have been studied extensively in their respective fields.
For example, BISG and its variants are well-discussed in the analysis of healthcare systems (where BISG was first developed), in fair lending audits \cite{voicu2018, cfpb2014}, in ensuring fairness for ads~\cite{meta-ads2023}, and in evaluating congressional redistricting proposals~\cite{deluca-curiel2023}. Using secure two-party computation tools such as commutative encryption and homomorphic encryption for performing joins and computing on associated features have been widely studied (\cite{PSWW18, IKNPRSSSY19, BKMS20} to name a few) and local DP has also been comprehensively studied (\cite{EvfimievskiGeSr03, Warner65, KasiviswanathanLeNiRaSm11} to name a few).

AirBnB's Project Lighthouse~\cite{LighthouseAirbnb_2020} is a relevant precedent for PPRE. 
The aim of Project Lighthouse was to enable measurements of disparities in guests' booking acceptance rates across perceived race/ethnicity groups in an anonymized manner. Lighthouse shares some of the same goals with PPRE, namely enabling fairness analytics while assuring member privacy. They however make different choices for the privacy-enhancing technologies and the method for estimating demographics.
In particular, Lighthouse uses an external partner to facilitate the computation, applying asymmetric encryption in the data exchange with the partner and syntactic disclosure control methods ($k$-anonymity and $p$-sensitivity) to safeguard the privacy of members present in data tables. PPRE relies on differential privacy and secure two-party computation without an external facilitator instead. Lighthouse also relies on a third-party research partner for the demographic estimation function while PPRE uses BISG for this. 

PPRE shares some elements in common with the Variance Reduction System (VRS) for measuring fairness in relation to race/ethnicity across products and improving fairness in ad delivery~\cite{meta-ads2023, meta-ppre}. Specifically, VRS uses BISG for estimating race/ethnicity demographics. 
VRS also applies differential privacy on aggregated measurement to mask the members' inclusion in the ad-impressed group. In contrast, PPRE uses DP to mask members' inclusion in the self-ID survey sample set. The fairness system, similar to PPRE, also uses secure multiparty computation and differential privacy to make the system privacy-preserving. The difference being that they, similar to Project Lighthouse's approach, rely on external facilitators to participate in the multi-party protocol to collect encrypted survey responses and add DP to the output at the end of the output. 
In our case, we build a two-party secure computation protocol across the different entities within LinkedIn without relying on external facilitators and add DP to the input data right at the beginning of the computation. 

\paragraph{Structure of the Paper.} 
The rest of the paper flows as follows.
The next two sections (\sectionref{f-audits}-\sectionref{pets}) give details on the key components of the PPRE privatized fairness measurement system.
We start with elements needed for generic fairness measurement for race/ethnicity in \sectionref{f-audits}.
These include BISG, the generator for the race/ethnicity demographic data, and the disparity estimator functions, processing functions for producing the disparity estimator results.
Then we examine the privacy-enhancing technologies (PETs) that drive PPRE in \sectionref{pets}.
\sectionref{ppre-system} describes how we stitched all these pieces together.
\sectionref{impl} summarizes key implementation details and discusses a simple demonstration of the PPRE.
\sectionref{conc} summarizes our work and findings. We also highlight some ongoing or future lines of effort.

%% file: building-blocks-f-audits.tex
\section{Tools for Race \& Ethnicity Fairness Measurements}
\label{sec:f-audits}

In this work, we focus on the kinds of race and ethnicity fairness measurements that can be recast as disaggregated evaluations~\cite{barocas2021} of AI systems, i.e. assessments and comparisons of system performance on different race/ethnicity groups of people.
An example of AI fairness measurements includes predictive rate parity measurements sometimes used at  LinkedIn~\cite{quinonero2023}.
The last step of PPRE is the computation of those fairness measurements using what can be called \emph{disparity estimators}~\cite{chen2019, elzayn2023}.

To generate fairness measurements for race and ethnicity, we need two data elements: member-level data on race and ethnicity and member-level data on system performance. 
For AI systems, member-level data on performance will typically follow the schema: (member identifier, ground truth data, AI output). 
This kind of data can then be used to compute AI performance statistics like classification accuracy, AUC, error rates, etc.
Disparity estimators are responsible for estimating group-wise sample statistics by aggregating over the member-level data elements. 

An example of an disparity estimator for AI models is the standard false positive rate (FPR) disparity estimator for classification models.
Assume $Y$ is the ground truth classification, $\hat Y$ is the model's predicted classification, and $R$ is the race/ethnicity categorical indicator taking values $R=j, j \in \{1, \cdots, k\}$.
This gives a member-level data set for the audit: $\{R_i, Y_i, \hat Y_i \}_i$.
Then the FPR disparity estimator is the following group-wise sample mean statistic:
\begin{equation}
    \mu[j] = \sum_i^n \frac{\mathbb{I}(R_i=j) \cdot \mathbb{I}(\hat Y_i=1, Y_i=0)}{\sum_i^n \mathbb{I}(R_i=j)}, \; \forall j \in \{1, \ldots, k\}
    \label{eq:sample-mean}
\end{equation}
where $\mathbb{I}$ denotes the binary indicator function.
The $\mu[j]$ are statistical estimators for the probability of the classification model causing a false positive for each of the $k$ race/ethnicity categories.
This disparity estimator fits nicely when the audit demographic attribute is categorical (e.g. for gender).
An important PPRE design goal was to avoid assigning individual members to categorical race/ethnicity groups, just on principle.
And Elliott et al.~\cite{elliott2009} point out that, although assignment is procedurally simpler, it incurs more error. 
So we need a different kind of disparity estimator for fairness measurements using the BISG-based PPRE system.

The design of PPRE is motivated by the absence of robust and comprehensive member-level data on race/ethnicity.
We address this complication by generating probabilistic proxies for the member-level race/ethnicity attribute (details in \sectionref{bisg} below).
And, to accommodate the use of probabilistic proxies of race/ethnicity, we generalize standard disparity estimators to handle \emph{probabilistic} demographic attributes in place of the categorical attributes like in the example Equation \ref{eq:sample-mean} above (details in \sectionref{audit-func} below).

\subsection{BISG for Estimating Probabilistic Race/Ethnicity}
\label{sec:bisg}
\begin{figure}
    \centering
    \includegraphics[width=0.5\linewidth]{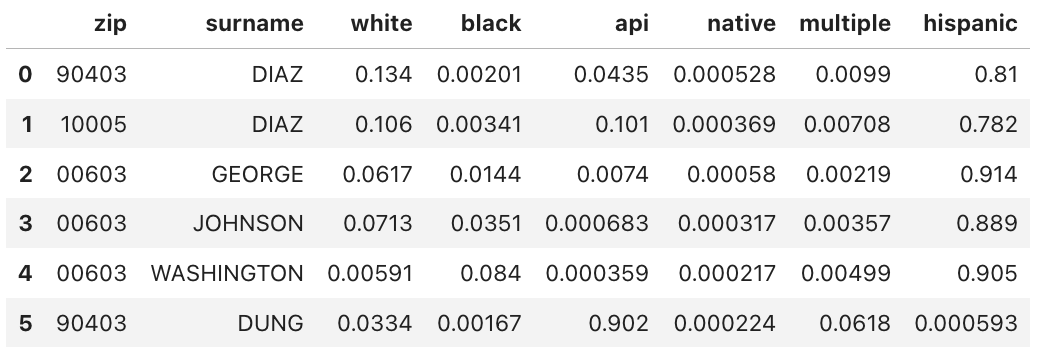}
    \caption{A Mock Sample of BISG Estimates.\\ \tiny{The rows do not depict real individuals recorded in a dataset. Any resemblance is purely coincidental.}}
    \label{fig:bisg-sample}
\end{figure}

We summarize the development of the Bayesian Improved Surname Geocoding (BISG) method from discussions in earlier papers~\cite{elliott2009, voicu2018, imai2022}.
The goal of BISG is to produce estimates of an individual's probability $R$ of belonging to one of six exclusive race/ethnicity categories (a $k=6$ dimensional categorical random variable). 
The BISG estimate is informed by two input variables, the member's surname $S$ and their location $G$, using a straightforward application of Bayes theorem.
The BISG estimate for an individual is in the form of a categorical posterior conditional probability $R \, | \, S, G \sim Categorical(\vec{p} \; | \, S, G)$.
The BISG model equation is just the conditional probability equation:
\begin{equation*}
    \mathsf{Pr}(r|s, g) = \frac{\mathsf{Pr}(r,s,g)}{\mathsf{Pr}(s,g)} = \frac{\mathsf{Pr}(r,s,g)}{\sum_{r=1}^{k=6} \mathsf{Pr}(r,s,g) } \label{eq:Bayes-SG}\;.
\end{equation*}
We make the following conditional independence assumption to simplify the model: $S \perp G| R$, i.e. surnames are independent of location given the member's race/ethnicity. 
The assumption simplifies the full joint probability equation:
\begin{align*}
    \mathsf{Pr}(r,s,g) &= \mathsf{Pr}(s) \cdot \mathsf{Pr}(r|s) \cdot \mathsf{Pr}(g|r,s) = \mathsf{Pr}(s) \cdot \mathsf{Pr}(r|s) \cdot \mathsf{Pr}(g|r) \\
    \mathsf{Pr}(r|s, g) &= \frac{\mathsf{Pr}(s) \cdot \mathsf{Pr}(r|s) \cdot \mathsf{Pr}(g|r)}{\sum_{r=1}^{k=6} \mathsf{Pr}(s) \cdot \mathsf{Pr}(r|s) \cdot \mathsf{Pr}(g|r) } \\
    \mathsf{Pr}(r|s, g) &= \frac{\mathsf{Pr}(r|s) \cdot \mathsf{Pr}(g|r)}{\sum_{r=1}^{k=6} \mathsf{Pr}(r|s) \cdot \mathsf{Pr}(g|r) } \;.
\end{align*}
To implement this Bayesian model, the posterior equation above $\mathsf{Pr}(r|s, g)$ needs real-world tabular data to inform the conditional probabilities of $G|R$ and $R|S$. 
Practitioners use frequency tables generated from the 2010 public U.S. Census on ``Frequently Occurring Surnames'' and the ``Census Summary File'' survey summaries~\cite{census_surnames_2016, census_summary_2012} to populate the $R|S$ and the $G|R$ probabilities. 
Location information $G$ is aggregated to the level of ZIP Code Tabulation Areas (ZCTAs).
These Census-derived tables follow U.S. federal standards for a combined format for race and ethnicity categories~\cite{omb_1997} that consist of $6$ race \& ethnicity categories: White, Black, Hispanic or Latino (a combined category), American Indian or Alaska Native, Asian or Pacific Islander (a combined category). Figure \ref{fig:bisg-sample} depicts hypothetical results of BISG estimates for mock individuals.

Voicu~\cite{voicu2018} extends BISG to the Bayesian Improved First name Surname Geocoding (BIFSG) method by including first names as an input dimension, extending the conditional independence assumption, and using the following Bayes update equation:
\begin{align*}
    \mathsf{Pr}(r|f, s, g) &= \frac{\mathsf{Pr}(r|s) \cdot \mathsf{Pr}(g|r) \cdot \mathsf{Pr}(f|r)}{\sum_{r=1}^{k=6} \mathsf{Pr}(r|s) \cdot \mathsf{Pr}(g|r) \cdot \mathsf{Pr}(f|r) } \;. 
\end{align*}
This variant uses an secondary survey of first names and race/ethnicity to inform the $\mathsf{Pr}(f|r)$ probabilities.
This survey was not run by the U.S. Census.
We rely on only the public Census tables for our implementations.

The conditional independence assumption $S \perp G| R$ is a useful simplification. 
It reduces the data needs for the BISG method. 
The assumption is likely untrue in the strictest sense. 
This assumption would be violated if for a given race/ethnicity group, there are correlations between last name and location. 
For example, Asian Americans with specific names (e.g. Kim) having a higher presence in some locations (e.g. Korea Town in Los Angeles) would be a violation.
This pattern of correlation between names and location is common enough as Imai et al.~\cite{imai2022} note.
Future work may explore methods for addressing this violated assumption.

Besides violations of theoretical assumptions, researchers have documented other practical shortcomings in BISG~\cite{rieke2022imperfect, chen2019, kallus2022, pai-demog2021}.
BISG estimates are reported to be worse performing for younger age groups, possibly due to the age of the Census survey powering the method (2010 Decennial Survey)~\cite{rieke2022imperfect}.
Disparity estimators based on BISG estimates or proxies will have statistical bias under certain conditions~\cite{chen2019, kallus2022}.
And researchers have argued that BISG's race/ethnicity categorization are too narrow to properly highlight groups that experience distinct disparities~\cite{pai-demog2021}. 
Without opining on the categorization's optimality, we use this categorization because it is the most widespread standard used in the U.S. government as well as in academia. 
Future work is needed to refine the BISG method for LinkedIn member populations as well to better address the statistical limitations of BISG-based disparity estimators.

\subsection{Disparity Estimators}
\label{sec:audit-func}
We need disparity estimators that are not based on binary or categorical group membership unlike standard disparity estimators.
Our default approach is to apply ideas from Chen et al.~\cite{chen2019} by generalizing the sample mean disparity estimators to accommodate our use of probabilistic race/ethnicity estimates instead of categorical race/ethnicity attributes.
To summarize briefly, we replace the hard indicator functions in Equation \ref{eq:sample-mean} with soft indicator functions which is modeled by the estimated race probabilities.
Chen et al. ~\cite{chen2019} and Elzayn et al.~\cite{elzayn2023} further discuss the asymptotic properties of these kinds of disparity estimators.

There are different disparity estimators that may apply for fairness measurements on AI and product systems.
We discuss only three out of many possible kinds below and highlight scenarios where they may be relevant.
\begin{enumerate}
    \item \textbf{Model Performance Disparity Estimator}. These are the standard \emph{AI model fairness measurements}. Given the model performance metric $f(Y, \hat Y)$, a function of the ground truth and the model prediction, the probabilistic disparity estimator generalizes Equation \ref{eq:sample-mean} into:
    \begin{equation}
        \mu[j] = \sum_i^n \frac{\mathsf{Pr}(R_i=j) \cdot f(Y_i, \hat Y_i)}{\sum_i^n \mathsf{Pr}(R_i=j)}, \; \forall j \in \{1, \ldots, k\}
        \label{eq:soft-ai-sample-mean}
    \end{equation}
    \item \textbf{Output Metric Disparity Estimator}. These are disaggregated evaluations for outcome metrics. This type of audit is more applicable \emph{outcome equity audit} or \emph{product equity audit}. E.g. a researcher wants to estimate what fraction of the different race/ethnicity groups hold senior roles in the LinkedIn network. This kind of evaluation differs from the previous function in that it just generates group-wise sample means on a scalar signal or column $Y$ instead of on an input-output comparison $f(Y, \hat Y)$.
    \begin{equation}
        \mu[j] = \sum_i^n \frac{\mathsf{Pr}(R_i=j) \cdot Y_i}{\sum_i^n \mathsf{Pr}(R_i=j)}, \; \forall j \in \{1, \ldots, k\}
        \label{eq:soft-sample-mean}
    \end{equation}
    \item \textbf{Probabilistic Count Disparity Estimator}. These are \emph{inclusiveness measurements} that aim to estimate or check if there is at least a threshold number $T$ of members from a race/ethnicity group in a member pool. The use of probabilistic race/ethnicity estimates makes this necessarily a probabilistic check. If the required inclusiveness count threshold is $T=1$ for race/ethnicity group $j$, then the disparity estimator is the following probability estimator:
    \begin{align}
        \mathsf{P}_{Equity} &= \mathsf{Pr}(N_{j} \geq 1 | N_{pool}) = 1 - \mathsf{Pr}(N_{j} = 0 | N_{pool}) \notag \\
        \mathsf{P}_{Equity} &= 1 - \prod_i^{N_{pool}}\Big(1 - \mathsf{Pr}(R_i = j)\Big) \;. 
        \label{eq:soft-count}
    \end{align}
    The auditing party could then compare the estimated probability to a specified probability level e.g. $\mathsf{P}_{Equity} \gtrless 0.90$ to say with $90\%$ certainty if the pool includes at least $1$ member from the race/ethnicity group of interest. If the inclusiveness count threshold is $T>1$, then $N_{j} \sim PoiBin$ which requires more detailed calculations.
\end{enumerate}

The last step of the PPRE workflow (more on this in \sectionref{ppre-system}) makes use of these kinds of estimators.
The first two estimators are linear functions of system test values, $f(Y_i, \hat Y_i) \; \& \; Y_i$ respectively. The linearity property is important for the homomorphic encryption scheme.
The third estimator is a function of only race and ethnicity probabilities which are known and visible to the measuring or auditing party (P1). 
Inclusion in the sample pool is the only signal coming from the party or system to be measured (P2).
So the third estimator is also feasible in the P2 homomorphic encryption scheme discussed in the next section.

%% file: building-blocks-privacy.tex
\section{Privacy-Enhancing Technologies}\label{sec:pets}
We build our PPRE system with the aid of privacy enhancing technologies like secure two-party computation and differential privacy. We explain them in more detail below. Throughout this paper, we use $\secr$ to denote the computational security parameter.

\subsection{Secure Two-Party Computation}
Secure two-party computation is an advanced cryptographic primitive that allows two parties with private datasets to engage in a protocol, exchanging messages back and forth, to compute any function on their joint inputs without revealing any information to each other beyond what is revealed by just the function’s output. The protocol, at a high level, can be thought of as exchanging “encrypted” versions of their inputs (that are not decryptable by the other party), and “computing over this encrypted data". In particular, this primitive focuses on ensuring (a) privacy of the input datasets, and (b) privacy of the intermediate steps in the computation. Formally, privacy is guaranteed through a "real/ideal" world paradigm where: the real world denotes the actual protocol execution; the ideal world consists of a trusted third party that collects both parties' inputs, computes the function and returns the output; and security is established by comparing the outcome of an adversary participating in the real protocol execution with its participation in the ideal world and proving that these two are indistinguishable. This primitive has been studied extensively, starting with the work of ~\cite{Yao82}. We refer to \cite{Lindell21} for a more detailed study of this topic.

In this work, we build a secure two-party computation protocol to compute a specific function (tailored to performing a fairness measurement test) that relies on techniques from two cryptographic primitives: commutative encryption and additive homomorphic encryption.

\subsubsection{Commutative Encryption}
Commutative encryption, as the name suggests, has the property that for any message, given two keys, the message can be encrypted with those keys one after the other, in any order, to compute the same resulting ``doubly-encrypted" ciphertext. Subsequently, this can then be decrypted using the two keys, one after the other, in any order as well. 

Formally, a commutative encryption scheme $\com = (\comkeygen, \comenc, \comdec)$ is a symmetric encryption scheme (with a deterministic encryption algorithm) over message space $\M$ with the following structure:
\begin{itemize}
\item $\sk \gets \comkeygen(1^\secr)$
\item $\ct \gets \comenc_\sk(m)$
\item $m / \bot \gets \comdec_\sk( \ct)$
\item For any two keys $\sk_1, \sk_2$ and any message $m \in \M$, 
	$$\comenc_{\sk_1}(\comenc_{\sk_2}(m)) = \comenc_{\sk_2}(\comenc_{\sk_1}(m))$$
\end{itemize}

A popular example is the Pohlig-Hellman cipher in which the message space $\M$ is any Elliptic Curve group $\G$ in which the Decisional Diffie Hellman (DDH) assumption holds. Note that in the presence of a random oracle $\RO$, a message in any arbitrary message space $\M$ can be transformed into an element of group $\G$. In \sectionref{impl}, we discuss more details on how we instantiate the random oracle and which group we use in our protocol.

Commutative encryption (when instantiated with a fresh $\RO$ instance in each protocol execution), serves as a building block in constructing private set intersection (PSI) protocols~\cite{Meadows86,HFH99} that allow two parties to engage in a secure computation protocol and learn only the intersection of their sets. We use the same methodology in our custom two-party computation protocol and detail that in \sectionref{ppre-system}. 

\subsubsection{Homomorphic Encryption}
Homomorphic encryption~\cite{Paillier99, Gen09} is a public key encryption scheme in which, without knowing the secret key, given
two ciphertexts encrypted with the same public key, one can compute a new ciphertext that is an encryption of any function applied to the underlying messages. A specific version of it is called additive homomorphic encryption which enables computing additions (and arbitrary linear combinations) of the underlying messages.

Formally, an additive homomorphic encryption scheme is a public-key encryption scheme $\ahe=(\ahekeygen,\aheenc,\aheadd,\ahedec)$ over message space $\M$ with the following structure: 
\begin{itemize}
\item $(\pk, \sk) \gets \ahekeygen(1^\secr)$
\item $\ct \gets \aheenc_\pk(m;r)$
\item $m / \bot \gets \ahedec_\sk( \ct)$
\item Homomorphic addition: $\ahedec_\sk(\aheadd(\aheenc_\pk(m_1), \aheenc_\pk(m_2))) = (m_1+m_2)$  $\forall m_1, m_2\in \M$.
\item Homomorphic multiplication with constant: $\ahedec_\sk(\aheadd(c, \aheenc_\pk(m))) = (c\cdot m)$  $\forall c, m \in \M$.
\end{itemize}
In the notation above, we loosely overload the function $\aheadd$ to denote both homomorphic addition and multiplication by a constant. Further, we implicitly assume that each homomorphic evaluation is followed by a refresh operation, where the resulting ciphertext is added with an independently generated encryption of zero. This is required in our protocols to ensure that the randomness of the final ciphertext is independent of the randomness used in the original set of ciphertexts. For popular additively homomorphic encryption schemes such as Paillier encryption \cite{Paillier99} (based on the Decisional Composite Residuosity assumption), a homomorphically evaluated ciphertext is statistically identical to a fresh ciphertext. We refer to~\cite{Paillier99} for formal definitions of correctness and CPA security.

Apart from the above, we also use traditional randomized symmetric encryption (denoted by the tuple of algorithms $(\symkeygen, \symenc, \symdec)$ in our solution.

\subsection{Differential Privacy}
Differential Privacy (DP) \cite{DworkMcNiSm06}, is a technique that adds carefully curated “noise” or randomness to ensure that an adversary cannot determine with certainty whether an individual's data was used or not in the computation.  We will be focusing on the \emph{local model} of DP, where randomness is applied to each individual's data prior to any aggregation, rather than the \emph{global model} of DP that allows for collecting data and then applying randomness as part of any aggregation.  The local model provides more privacy protections than the global model, but at the cost of lower utility.  

Someone looking at a dataset that has local DP applied to each record will not know which records are true and which ones have been modified. The challenge is to find the right amount of noise to add to preserve the privacy of each user while keeping some fidelity in the dataset. Every DP algorithm is parameterized by $\epsilon$ which quantifies the privacy risk - larger $\epsilon$ means the privacy risk is higher and hence increases the chance that any single record had not been falsified. Formally, an algorithm $\mathcal{M}: \mathcal{X} \rightarrow \mathcal{Y}$ is $\epsilon$-locally DP \cite{EvfimievskiGeSr03, Warner65, KasiviswanathanLeNiRaSm11} if for all possible inputs $x, x' \in \mathcal{X}$ and for all outcomes $\cS \subseteq \mathcal{Y}$, we have $\mathsf{Pr}[\mathcal{M}(x) \in \cS] \leq e^\epsilon \mathsf{Pr}[\mathcal{M}(x') \in \cS]$.

In our work, we rely on the traditional local DP algorithm called randomized response~\cite{Warner65}, which works as follows: for every record in the dataset, we flip the record to take on a different value with a fixed probability (that is determined by $\epsilon$). In more detail, suppose we're interested in applying local DP to records that can take on one of $k$ values $\{1,,\ldots,k\}$. The randomized response mechanism $\mathcal{M}: \{1,,\ldots,k\} \rightarrow \{1,,\ldots,k\}$, on any input $i$, retains the value $i$ as is with probability $\frac{e^\epsilon}{e^\epsilon + k - 1}$ and flips it to each of the other $(k-1)$ entries with probability $\frac{1}{e^\epsilon + k - 1}$, i.e.
\begin{align*}
\mathsf{Pr}[\mathcal{M}(i)  = i] &= \frac{e^\epsilon}{e^\epsilon + k - 1} 
\\
\mathsf{Pr}[\mathcal{M}(i)  \in \{ 1, \cdots k\} \setminus \{i\} ] &= \frac{k-1}{e^\epsilon + k - 1} .
\end{align*}

%% file: ppre-details.tex
\section{The PPRE System: Stitching It All Together}
\label{sec:ppre-system}
\begin{figure}[ht!]
\centering
\includegraphics[width=170mm, height=190mm]{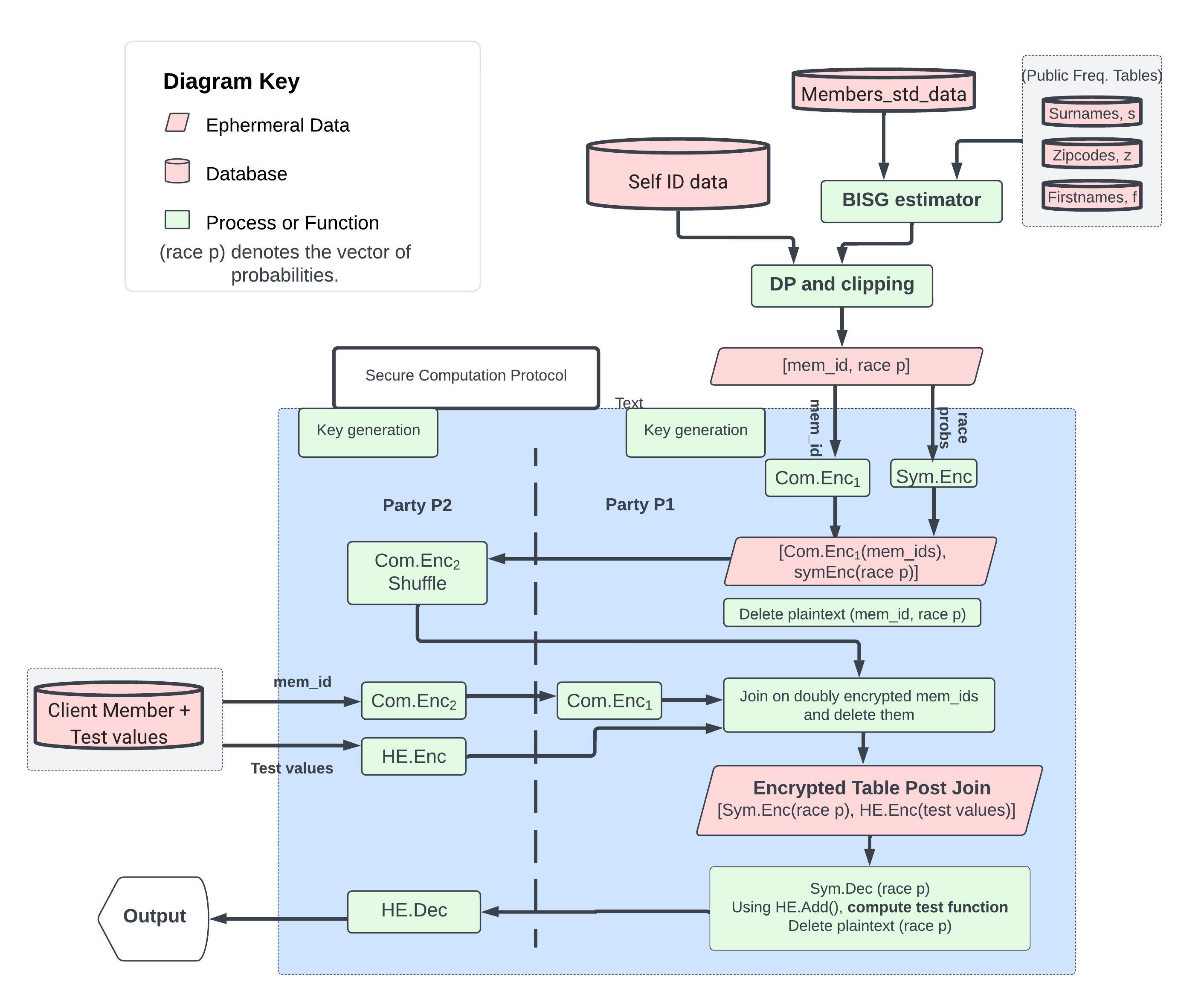}
\caption{PPRE system}
\label{fig:ppre-system}
\end{figure}

We consider two entities in our system – (a) the tester, which is the data privacy/responsible AI team and (b) the test client which is any product owner whose AI model we’d like to test. We denote the tester as party $P_1$ and the test client as party $P_2$.  Our overall system, pictorially, is presented in \figureref{ppre-system}. 

 At a high level, our system works as follows: $P_1$ first computes a privacy-preserving version of a probabilistic race/ethnicity demographic dataset. The two parties then engage in a secure two-party computation protocol to join both their private datasets and evaluate the test function to compute the resulting aggregated outputs. The test functions listed in \sectionref{audit-func} are weighted sample mean operations so they can be implemented via standard aggregation methods (e.g. in Spark/Scala). Many of these test functions are also linear operations over the client's test values and so, can be fully-supported with the additive homomorphic encryption (outlined in more detail below).

\subsection{Data Preparation and Local DP}
\label{sec:clipping}
To compute a privacy-preserving version of a probabilistic race/ethnicity demographic dataset, $P_1$ combines BISG with Self-ID datasets and applies local DP and clipping. We expand on this.

First, $P_1$ applies local DP (randomized response) to the Self-ID data. In more detail, each Self-ID record can be represented as a one-hot vector of $k$-dimension, so is of the form $(1,0,0,0,0,0)$ (if we consider the six race and ethnicity categories mentioned earlier) with $k=6$ where we assign $1$ corresponding to the race/ethnicity column that the member self-identified as. We then apply traditional randomized response to keep the bit $1$ as is with probability $\frac{e^\epsilon}{(k-1+e^\epsilon)}$ and otherwise flip it to $0$ and pick one of the other $(k-1)$ entries uniformly at random to flip from $0$ to $1$ giving an effective flip probability of $\frac{1}{(k-1+e^\epsilon)}$ for each of the $(k-1)$ entries. The parameter $\epsilon$ is tunable and we pick it depending on the application.

Then, $P_1$ computes the BISG probabilistic estimates and combines this with the above local DP version of the Self-ID table. For the entire set of records (BISG + local DP Self-ID), we apply probabilistic clipping to prevent the possibility of 100$\%$-certain assignment of members to any single race/ethnicity category. In more detail, when a race/ethnicity probability in a row exceeds a clipping threshold $\bT$, we reduce it by a small uniformly sampled random value to bring it below the threshold. Then, we increase the probability values for the other race/ethnicity categories by a properly-scaled random draw from a Dirichlet (vector) distribution. To determine $\bT$, in the entire set of BISG data, let $\cS$ denote the set of the largest probability values in each record. We set $\bT$ to be such that $90\%$ of the entries in $\cS$ are below $\bT$. The effect of this clipping is to prevent determining which records were computed with BISG and which ones were Self-ID records by looking at the intermediate steps or the outputs of this measurement flow. In the flow diagram, we use $\text{(race p)}$ to denote the vector of probability values $\racep$.

\subsection{Secure Computation}
\label{sec:secure-computation}
At the end of the above phase, party $P_1$ has a privacy-preserving version of race/ethnicity data which can be expressed as a table containing one column for the member id and 6 columns for the race/ethnicity probabilities. In short, we denote each row as a tuple of $(\memid, \racep)$. $P_2$ has as input tuples of $(\memid, \values)$ where $\values$ denotes the vector of values held by the test client for each member id. Let $\G$ denote the group over which the commutative encryption scheme $\comenc$ operates, $\RO: {0,1}^* \rightarrow \G$ denote a random oracle that maps arbitrary length strings to an element of group $\G$. The message space for the symmetric and additive HE schemes is integers (we scale up the float probability values to appropriate integer domains).

Our secure two-party computation protocol draws inspiration from literature in PSI-Sum~\cite{IKNPRSSSY19} and circuit PSI~\cite{PSWW18} to compute a privacy-preserving join and perform aggregate calculation. The protocol proceeds as follows:
\begin{itemize}
	\item \textbf{Key Generation:} In our actual system, this step occurs even before or in parallel to the data preparation phase but we call it out here for ease of exposition.
		\begin{itemize}
		\item Instantiate a fresh instance of the random oracle $\RO$.
		\item $P_1$ generates a key for the commutative encryption scheme $\sk_1 \leftarrow \comkeygen(1^\secr)$ and the symmetric encryption scheme $\symsk \leftarrow \symkeygen(1^\secr)$.
		\item $P_2$ generates a key for the commutative encryption scheme $\sk_2 \leftarrow \comkeygen(1^\secr)$ and a key pair for the $\ahe$ scheme $(\ahepk,\ahesk) \leftarrow \ahekeygen(1^\secr)$.
		\end{itemize}
	
	\item \textbf{Encrypt $P_1$ data:} $P_1$ does the following:  
	\begin{itemize} 
	\item Encrypt $\memid$ using the commutative encryption scheme and $\racep$ using the symmetric encryption scheme. 
	\item Send $\{(\comenc_{\sk_1}(\RO(\memid)), \symenc_\symsk(\racep; \rand_i))\}$ to $P_2$ where $\rand_i$ denotes the randomness used for the $i^{th}$ row.
	\item Then, $P_1$ deletes its plaintext input dataset $\{\memid, \racep\}$.
	\end{itemize}
	
	\item \textbf{Doubly encrypt $P_1$ data:} Using its commutative encryption key $\sk_2$, $P_2$ doubly encrypts $P_1$'s data to compute $\{(\comenc_{\sk_2}(\comenc_{\sk_1}(\RO(\memid))), \symenc_\symsk(\racep; \rand_i))\}$. $P_2$ shuffles this dataset and sends it back to P1. 	
	
	\item \textbf{Encrypt $P_2$ data:} $P_2$ does the following:  
	\begin{itemize} 
	\item Encrypt $\memid$ using the commutative encryption scheme and $\values$ using the additive HE scheme. 
	\item Send $\{(\comenc_{\sk_2}(\RO(\memid)), \aheenc_\ahepk(\values; \rand_j))\}$ to $P_2$ where $\rand_j$ denotes the randomness used for the $j^{th}$ row. 
	\end{itemize}
	
	\item \textbf{Privacy-preserving join:} $P_1$ does the following:
	\begin{itemize}
	\item First, using its commutative encryption key $\sk_1$, $P_1$ doubly encrypts $P_2$'s member identifiers to compute $\{(\comenc_{\sk_1}(\comenc_{\sk_2}(\RO(\memid))), \aheenc_\ahepk(\values; \rand_j))\}$.
	\item Then, using the commutative property of the encryption scheme, $P_1$ can join both of these datasets with the common join key as the doubly encrypted member identifiers -- that is, 
	$\comenc_{\sk_1}(\comenc_{\sk_2}(\RO(\memid_1))) == \comenc_{\sk_2}(\comenc_{\sk_1}(\RO(\memid_2)))$ if $\memid_1 = \memid_2$.
	\item $P_1$ then deletes all the doubly encrypted member identifiers and computes the resulting joined table  $\{(\symenc_\symsk(\racep; \rand_i), \aheenc_\ahepk(\values; \rand_j))\}$.
	\item Here, observe that, even though P1 deletes its plaintext values in the first step after encrypting its data, after doubly encrypting, the shuffling step that $P_2$ performs adds another layer of privacy protection. It helps prevent $P_1$ from being able to associate which specific row of its original input was part of the join and which gets discarded (even if it doesn't know the underlying deleted member identifier). 
	\end{itemize}
	
	\item \textbf{Test computation:} Party $P_1$ does the following to compute the test: 
	\begin{itemize}
	\item Using key $\symsk$, decrypt the race/ethnicity probabilities to compute\\
 $\{(\racep, \aheenc_\ahepk(\values; \rand_j))\} \leftarrow (\symdec_\symsk(\symenc_\symsk(\racep; \rand_i)),$ \\ $\aheenc_\ahepk(\values; \rand_j))$  
	\item As explained in \sectionref{audit-func}, the test functions we are interested in require performing only linear combinations of the $\{\values\}$ and $\{\racep\}$. Using repeated calls to $\aheadd$, we can compute an encryption of the aggregate output: $\{(\mathsf{race_i}, \aheenc_\ahepk(\out_i, r_i))\}_{i=1}^{6}$ where the tuple has 6 values, one for each race/ethnicity type we identified in the data preparation stage. This is sent to $P_2$.
	\end{itemize}

	\item \textbf{P2 learns output:} Using additive HE key $\ahesk$, $P_2$ can run the decryption algorithm $\ahedec$ to learn the final aggregate output. Observe that if we care about only $P_1$ learning the output and not $P_2$, we can apply a simple transformation where in the previous step, $P_1$ masks the encrypted aggregates with a private mask, $P_2$ sends back the decrypted masked outputs to $P_1$ (which doesn't reveal anything to $P_2$) and $P_1$ can remove the mask to learn the output.
\end{itemize}
Moreover, to further prevent re-identification of individuals from the output of the computation, we mandate a governance process to ensure that P2’s dataset contains a minimum threshold of records and that P2 doesn’t perform repeated measurements with just a few entries replaced/modified.

%% file: implementation.tex
\section{System Performance and Demonstration}
\label{sec:impl}
We first discuss details around how we implemented our system and then discuss a case study with some performance benchmarking. 

\paragraph{Parameters:}
We discuss the choices for the different cryptographic parameters in our system. First, we set the computational security parameter $\secr = 128$. For the commutative encryption scheme, we pick group $\G$ as the Elliptic Curve Curve25519~\cite{Ber06}. We instantiate the random oracle with cryptographic hash function SHA-256 and hash into the group $\G$ using rejection sampling. For the additive homomorphic encryption scheme, we use the Paillier cryptosystem~\cite{Paillier99} with a 2048 bit modulus and instantiate the symmetric encryption using AES with 256 bits as the key size. 

For local DP, we set $\epsilon = 4.5$, which leads to a flipping rate $=5.26\%$. Recall that, the clipping threshold $\bT$ is determined by looking at the set of the largest probability values in each record of the BISG data (as outlined in \sectionref{clipping}). Based on analysis, we set $\bT = 0.825$ by default.

\begin{figure}[ht!]
\centering
\includegraphics[width=180mm, height=100mm]{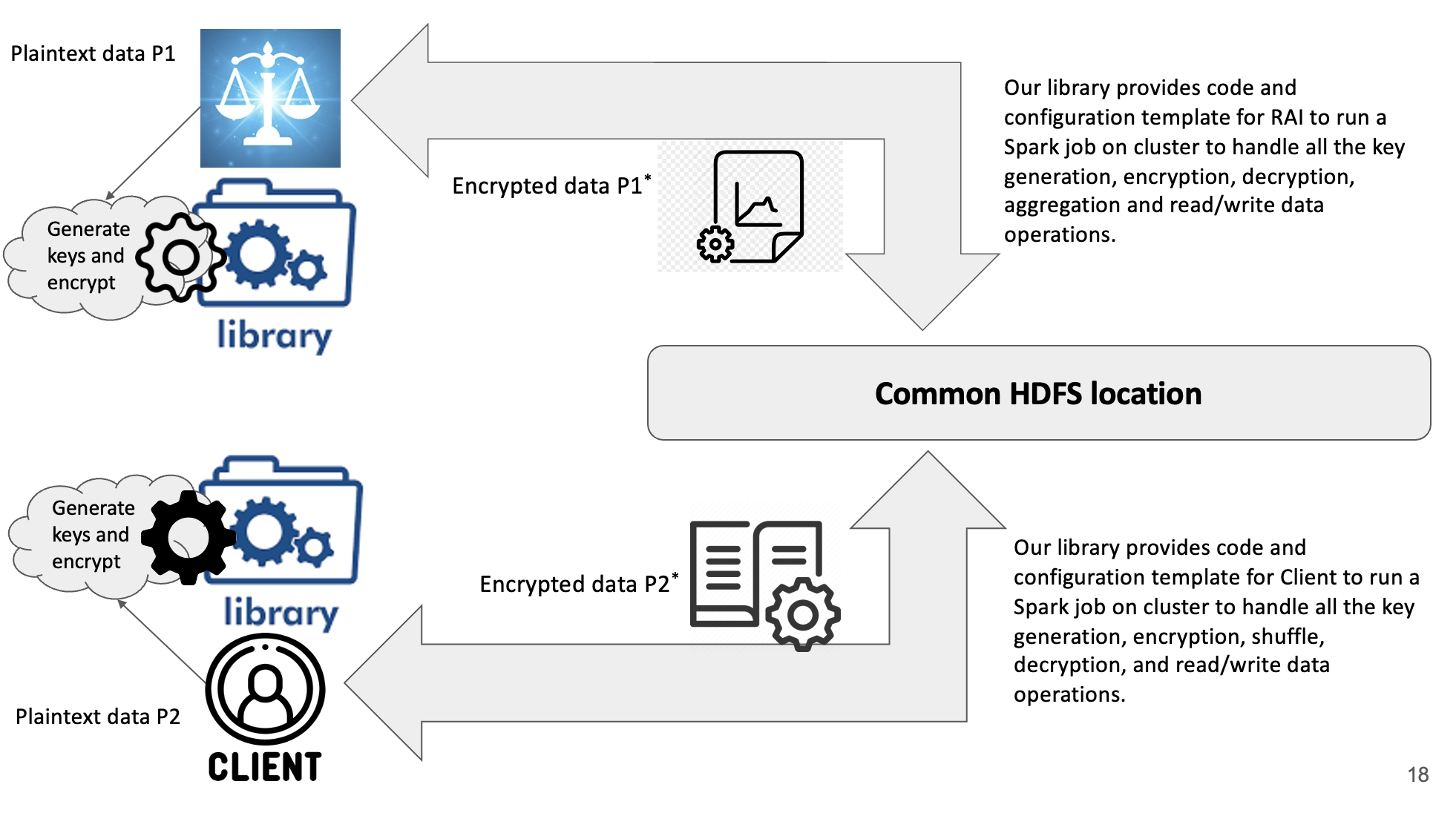}
\caption{PPRE data flows}
\label{fig:data-flows}
\end{figure}

Given the large volume of data that we conduct PPRE tests on and the fact that the applications typically do not need real-time flows, we do not implement ``direct" communication via a network between the two parties to exchange encrypted data. Instead, we enable data exchange by having each party store the output of their computation on the LinkedIn HDFS cluster and the receiving party reading data from the HDFS cluster (and subsequently deleting it). Only encrypted data will be stored at commonly shared locations on HDFS (without revealing the plaintext data or encryption keys across different parties). We leverage the Spark system within LinkedIn to implement the system and read/write large amounts of data through HDFS on top of LinkedIn infrastructure. Each party’s Spark job has a read-write lock scheme which can detect whether the other party has completed uploading their encrypted data before performing any subsequent processing. Also, we don't store any keys directly on HDFS or use any KMS service. Instead, the keys are only persisted in each party's Spark job memory during runtime and deleted right after the job finishes.
The diagram in \figureref{data-flows} illustrates the process in more detail. We use Spark’s native distributed mechanism and the flows are fully automated.

%% file: sample-analyses.tex
\paragraph{Benchmarking:}
We benchmarked our system using various settings. 
In general, the higher parallelism we set (e.g. increase a Spark cluster’s number of executors), the better the performance of the system. 
If we are conservative on the resource allocation: For a Spark cluster with 8GB driver memory, 4GB executor memory, 2 executor cores and 100 executors, our system can process datasets of sizes ranging from 1M*1M in around 8 mins to 20M*20M in under an hour.

\paragraph{Validation:}
We discuss some relevant modes of validation analyses for PPRE.
First, a validation of the performance of the baseline BISG on the LinkedIn U.S. member base. 
We use the cross entropy to compare BISG outputs with race/ethnicity responses from the Self-ID survey.
The Self-ID survey on a significantly self-selected subpopulation which means this is an imperfect comparison due to selection bias in the `ground truth` data.
We also do this comparison for a few other BISG variants that use some combination of first names, surnames, and zipcode for their outputs.

\begin{figure}
    \centering
    \includegraphics[width=0.65\linewidth]{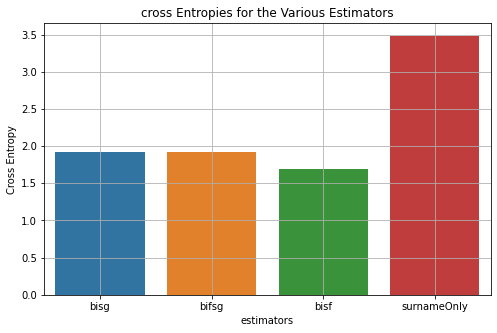}
    \caption{Comparing the cross entropy of BISG and its variants on a sub-sample of the LinkedIn Self-ID survey. Lower is better.}
    \label{fig:bisg-xent-cmp}
\end{figure}

Figure \ref{fig:bisg-xent-cmp} shows this comparison.
We find that the BIFSG method, which augments BISG with corrections for first names based on frequency tables from a Harvard survey~\cite{tzioumis_2018}, does not meaningfully change or improve the baseline BISG method.
While using only surnames and first names (BISF) yields a slightly lower cross-entropy (better performance metric), this marginal gain comes at the cost of diminished performance for smaller racial groups with lower base rates in the Self-ID survey.

Next, we examine sample fairness measurements from the PPRE method.
The analyses here \emph{forego the homomorphic encryption} part of PPRE to simplify numerical comparisons.
We test out two different PPRE-based measurements. The first is an output metric disparity measurement intended to estimate the fraction of White and Black people that hold senior roles in the LinkedIn network.
using the output metric disparity estimator (Equation \ref{eq:soft-sample-mean}) where $Y$ is the seniority indicator computed based on the members' profile entries.
We examine numbers from this measurement for the two race/ethnicity groups that use the least amount of synthetic aggregation, White and Black.
\begin{equation}
    \mu[j] = \sum_i^n \frac{\mathsf{Pr}(R_i=j) \cdot Y_i}{\sum_i^n \mathsf{Pr}(R_i=j)}, \; \forall j \in \{ \text{White}, \text{Black} \}
\end{equation}
Table \ref{tab:egri-bisg} shows the sample measurement. 
\begin{table}
    \centering
    \begin{tabular}{c|c}
    Race/Ethnicity Group & Estimated Seniority Share \\
    \hline
    White & 0.2334 \\
    Black/African & 0.2191 \\
    \end{tabular}
    \caption{PPRE-measurement comparing seniority shares across two race/ethnicity groups for a random sample of the LinkedIn U.S. population.} 
    \label{tab:egri-bisg}
\end{table}

These results are similar in relative order to results from a recent LinkedIn Economic Graph analysis on the same question~\cite{Baird2024} that used race/ethnicity responses from only the Self-ID survey ($W:0.3459 > B/A: 0.3337$).  
The estimated magnitudes of the disparity are not the same in both the PPRE and EGRI measurements. 
However the estimates are not directly comparable since they are on statistically dissimilar sample populations.

The second comparison uses the model performance disparity estimator in Eq. \ref{eq:soft-ai-sample-mean}.
We generate performance data for a hypothetical AI classification model and aim to calculate the race/ethnicity-based differences in the classification's false positive rate.
We use the following measurement function:
\begin{equation}
    \mu[j] = \sum_i^n \frac{\mathsf{Pr}(R_i=j) \cdot \mathbb{I}(\hat Y_i=1, Y_i=0)}{\sum_i^n \mathsf{Pr}(R_i=j)}, \; \forall j \in \{ \text{White}, \text{Black} \}
\end{equation}
where $Y$ is the ground truth classification and $\hat Y$ is the model's predicted classification.
Table \ref{tab:ero-bisg} shows an example of this evaluation. 
\begin{table}
    \centering
    \begin{tabular}{c|c}
        Race/Ethnicity Group & Estimated FPR \\
        \hline
        White & 0.00121 \\ 
        Black/African  & 0.00126 \\ 
    \end{tabular}
    \caption{PPRE measurements comparing group-wise false positive rates for a hypothetical classification model.}
    \label{tab:ero-bisg}
\end{table}

%% file: conclusion.tex
\section{Conclusion}
\label{sec:conc} 

In summary, our paper presents the PPRE system designed to enable useful fairness measurements based on race/ethnicity for U.S. LinkedIn members, in a privacy-preserving manner. Our investment in this technology will serve as the foundation to execute on our commitment to ensure that we are providing fair and equitable experiences that help all of our members and customers be more productive and successful. 

\subsection{Future Work}
There is further work to do on expanding the scope of our system to perform race/ethnicity based fairness measurements for all AI models at LinkedIn and enhance the privacy guarantees of our system. 
In particular, this includes the following from the privacy perspective:
\begin{itemize}
	\item Implementing more expressive measurement functions and expanding our system with other advanced secure computation protocols to perform those computations;
	\item Keeping the privacy-preserved version of individual race/ethnicity probabilities encrypted, even during the final calculation of the disparity estimator; \&
	\item Correcting the final output to aim for statistical accuracy by de-biasing the effect of adding local DP.
\end{itemize}
We also need to improve the fairness measurements themselves by improving the following:
\begin{itemize}
     \item Improving the calibration of BISG or its variants for the LinkedIn population;
     \item Developing more principled uncertainty quantification methods for BISG-based fairness measurements;
     \item Deriving disparity estimators with better statistical properties (e.g. better efficiency) for probabilistic demographic proxies. 
\end{itemize}

\subsection{Acknowledgements}
This work was the result of a fruitful and massive collaboration between the data privacy, responsible AI and legal teams. We would like to thank Joaquin Quiñonero Candela, Daniel Olmedilla, Souvik Ghosh, Ya Xu for their collaboration and leadership support; Daniel Tweed-Kent, Matthew Baird, Sam Gong, Adrian Rivera Cardoso, Katherine Vaiente, and Igor Perisic for their feedback. Finally, we would like to express our deep gratitude to those members who have trusted us with their Self-ID data. This work would not have been possible without your contribution.

%% file: refs.bib
@misc{LawitXu_2023, 
    author={Lawit, Blake and Xu, Ya}, 
    title={{Sharing LinkedIn’s Responsible AI Principles}}, 
    howpublished={\url{https://www.linkedin.com/blog/member/trust-and-safety/responsible-ai-principles}}, 
    year={2023}, month={Feb} 
}

@inproceedings{quinonero2023,
  title={{Disentangling and Operationalizing AI fairness at Linkedin}},
  author={Qui{\~n}onero Candela, Joaquin and Wu, Yuwen and Hsu, Brian and Jain, Sakshi and Ramos, Jennifer and Adams, Jon and Hallman, Robert and Basu, Kinjal},
  booktitle={Proceedings of the 2023 ACM Conference on Fairness, Accountability, and Transparency},
  pages={1213--1228},
  year={2023}
}

@inproceedings{Gen09,
  author       = {Craig Gentry},
  title        = {Fully homomorphic encryption using ideal lattices},
  booktitle    = {Proceedings of the 41st Annual {ACM} Symposium on Theory of Computing,
                  {STOC}},
  year         = {2009}
}

@misc{Baird2024, 
    title={US Race/Ethnicity Work Trends: Leadership and Remote Work}, 
    howpublished={\url{https://economicgraph.linkedin.com/content/dam/me/economicgraph/en-us/PDF/us-race-and-ethnicity-work-trends.pdf}}, 
    journal={LinkedIn Economic Graph}, 
    author={Baird, Matthew and Kavanagh-Smith, Danielle}, 
    year={2024}, month={Feb}
}

@inproceedings{barocas2021,
  title={Designing disaggregated evaluations of {AI} systems: Choices, considerations, and tradeoffs},
  author={Barocas, Solon and Guo, Anhong and Kamar, Ece and Krones, Jacquelyn and Morris, Meredith Ringel and Vaughan, Jennifer Wortman and Wadsworth, W Duncan and Wallach, Hanna},
  booktitle={Proceedings of the 2021 AAAI/ACM Conference on AI, Ethics, and Society},
  pages={368--378},
  year={2021}
}

@misc{LighthouseAirbnb_2020, 
    title={Measuring Discrimination on the {Airbnb} Platform}, 
    howpublished={\url{https://news.airbnb.com/measuring-discrimination-on-the-airbnb-platform/}}, 
    journal={Airbnb Newsroom}, 
    author={Airbnb}, year={2020}, month={Jun} 
}

@misc{hai-wrkshp2023, 
    author={Harvard School of Engineering and Applied Sciences},
    title={{How Can Bias Be Removed from Artificial Intelligence-Powered Hiring Platforms? Harvard-led institute to pursue fairness in online systems}}, 
    howpublished={\url{https://seas.harvard.edu/news/2023/06/how-can-bias-be-removed-artificial-intelligence-powered-hiring-platforms}}, 
    year={2023}, 
    month={Jun} 
}

@inproceedings{chen2019,
  title={Fairness under unawareness: Assessing disparity when protected class is unobserved},
  author={Chen, Jiahao and Kallus, Nathan and Mao, Xiaojie and Svacha, Geoffry and Udell, Madeleine},
  booktitle={Proceedings of the conference on fairness, accountability, and transparency},
  pages={339--348},
  year={2019}
}

@book{elzayn2023,
  title={Measuring and mitigating racial disparities in tax audits},
  author={Elzayn, Hadi and Smith, Evelyn and Hertz, Thomas and Ramesh, Arun and Goldin, Jacob and Ho, Daniel E and Fisher, Robin},
  year={2023},
  publisher={Stanford Institute for Economic Policy Research (SIEPR)}
}

@inproceedings{rieke2022imperfect,
  title={Imperfect Inferences: A Practical Assessment},
  author={Rieke, Aaron and Southerland, Vincent and Svirsky, Dan and Hsu, Mingwei},
  booktitle={Proceedings of the 2022 ACM Conference on Fairness, Accountability, and Transparency},
  pages={767--777},
  year={2022}
}

@article{kallus2022,
  title={Assessing algorithmic fairness with unobserved protected class using data combination},
  author={Kallus, Nathan and Mao, Xiaojie and Zhou, Angela},
  journal={Management Science},
  volume={68},
  number={3},
  pages={1959--1981},
  year={2022},
  publisher={INFORMS}
}

@article{elliott2009,
  title={Using the Census Bureau’s surname list to improve estimates of race/ethnicity and associated disparities},
  author={Elliott, Marc N and Morrison, Peter A and Fremont, Allen and McCaffrey, Daniel F and Pantoja, Philip and Lurie, Nicole},
  journal={Health Services and Outcomes Research Methodology},
  volume={9},
  pages={69--83},
  year={2009},
  publisher={Springer}
}

@article{voicu2018,
  title={Using first name information to improve race and ethnicity classification},
  author={Voicu, Ioan},
  journal={Statistics and Public Policy},
  volume={5},
  number={1},
  pages={1--13},
  year={2018},
  publisher={Taylor \& Francis}
}

@article{imai2022,
  title={Addressing census data problems in race imputation via fully Bayesian Improved Surname Geocoding and name supplements},
  author={Imai, Kosuke and Olivella, Santiago and Rosenman, Evan TR},
  journal={Science Advances},
  volume={8},
  number={49},
  pages={eadc9824},
  year={2022},
  publisher={American Association for the Advancement of Science}
}

@misc{cfpb2014, 
    title={Using publicly available information to proxy for unidentified race and ethnicity: A methodology and assessment}, 
    howpublished={\url{https://files.consumerfinance.gov/f/201409_cfpb_report_proxy-methodology.pdf}}, 
    journal={Consumer Financial Protection Bureau}, 
    author={Consumer Financial Protection Bureau}, 
    year={2014} 
}

@misc{census_summary_2012, 
    title={Summary File 1 Dataset}, 
    howpublished={\url{https://www.census.gov/data/datasets/2010/dec/summary-file-1.html}}, 
    journal={Census.gov}, 
    author={US Census Bureau}, 
    year={2011}, month={Jun} 
}

@misc{census_surnames_2016,
    title={Frequently Occurring Surnames from the 2010 Census}, 
    howpublished={\url{https://www.census.gov/topics/population/genealogy/data/2010_surnames.html}}, 
    journal={Census.gov}, 
    author={US Census Bureau}, 
    year={2016}, month={Feb} 
}

@misc{omb_1997, 
    title={Revisions to the Standards for the Classification of Federal Data on Race and Ethnicity}, 
    howpublished={\url{https://www.federalregister.gov/documents/1997/10/30/97-28653/revisions-to-the-standards-for-the-classification-of-federal-data-on-race-and-ethnicity}}, 
    journal={Federal Register}, 
    author={Office of Management and Budget}, year={1997}, month={Oct} 
}

@misc{tzioumis_2018,
    author = {Tzioumis, Konstantinos},
    publisher = {Harvard Dataverse},
    title = {{Data for: Demographic aspects of first names}},
    UNF = {UNF:6:5PcFwvADtKPydVpPOelYPg==},
    year = {2018},
    version = {V1},
    doi = {10.7910/DVN/TYJKEZ},
    howpublished={\url{https://doi.org/10.7910/DVN/TYJKEZ}}
}

@inproceedings{meta-ads2023,
  title={Towards Fairness in Personalized Ads Using Impression Variance Aware Reinforcement Learning},
  author={Timmaraju, Aditya Srinivas and Mashayekhi, Mehdi and Chen, Mingliang and Zeng, Qi and Fettes, Quintin and Cheung, Wesley and Xiao, Yihan and Kannadasan, Manojkumar Rangasamy and Tripathi, Pushkar and Gahagan, Sean and others},
  booktitle={Proceedings of the 29th ACM SIGKDD Conference on Knowledge Discovery and Data Mining},
  pages={4937--4947},
  year={2023}
}

@article{deluca-curiel2023,
  title={Validating the applicability of bayesian inference with surname and geocoding to congressional redistricting},
  author={DeLuca, Kevin and Curiel, John A},
  journal={Political Analysis},
  volume={31},
  number={3},
  pages={465--471},
  year={2023},
  publisher={Cambridge University Press}
}

@inproceedings{DworkMcNiSm06,
    author = {Dwork, Cynthia and McSherry, Frank and Nissim, Kobbi and Smith, Adam},
    title = {Calibrating noise to sensitivity in private data analysis},
    year = {2006},
    isbn = {3540327312},
    publisher = {Springer-Verlag},
    address = {Berlin, Heidelberg},
    url = {https://doi.org/10.1007/11681878_14},
    doi = {10.1007/11681878_14},
    booktitle = {Proceedings of the Third Conference on Theory of Cryptography},
    pages = {265–284},
    numpages = {20},
    location = {New York, NY},
    series = {TCC'06}
}

@inproceedings{EvfimievskiGeSr03,
author = {Evfimievski, Alexandre and Gehrke, Johannes and Srikant, Ramakrishnan},
title = {Limiting privacy breaches in privacy preserving data mining},
year = {2003},
isbn = {1581136706},
publisher = {Association for Computing Machinery},
address = {New York, NY, USA},
url = {https://doi.org/10.1145/773153.773174},
doi = {10.1145/773153.773174},
booktitle = {Proceedings of the Twenty-Second ACM SIGMOD-SIGACT-SIGART Symposium on Principles of Database Systems},
pages = {211–222},
numpages = {12},
location = {San Diego, California},
series = {PODS '03}
}

@article{KasiviswanathanLeNiRaSm11,
author = {Shiva Prasad Kasiviswanathan and Homin K. Lee and Kobbi Nissim
and Sofya Raskhodnikova and Adam Smith},
title = {What can we learn privately?},
year = {2011},
journal = {SIAM Journal on Computing},
volume = 40,
number = 3,
pages = {793--826},
url = {http://arxiv.org/abs/0803.0924}
}

@article{Warner65,
  author = {Warner, Stanley L},	
  title={Randomized Response: A Survey Technique for Eliminating Evasive Answer Bias},
  journal   = {Journal of the American Statistical Association 60},
  year={1965},
  url = {https://doi.org/10.2307/2283137.}
}

@inproceedings{Yao82,
  author       = {Andrew Chi{-}Chih Yao},
  title        = {Protocols for Secure Computations (Extended Abstract)},
  booktitle    = {23rd Annual Symposium on Foundations of Computer Science (FOCS)},
  year         = {1982},
  url          = {https://doi.org/10.1109/SFCS.1982.38}
}

@article{Lindell21,
  author       = {Yehuda Lindell},
  title        = {Secure multiparty computation},
  journal      = {Commun. {ACM}},
  volume       = {64},
  number       = {1},
  pages        = {86--96},
  year         = {2021},
  url          = {https://doi.org/10.1145/3387108},
}

@inproceedings{Paillier99,
  author    = {Pascal Paillier},
  title     = {Public-Key Cryptosystems Based on Composite Degree Residuosity Classes},
  booktitle = {EUROCRYPT},
  year      = {1999}
}

@INPROCEEDINGS{Meadows86,
	author={C. Meadows},
	booktitle={IEEE S \& P},
	title={A More Efficient Cryptographic Matchmaking Protocol for Use in the Absence of a Continuously Available Third Party},
	year={1986}
}

@inproceedings{HFH99,
  author    = {Bernardo A. Huberman and
               Matthew K. Franklin and
               Tad Hogg},
  title     = {Enhancing privacy and trust in electronic communities},
  booktitle = {ACM Conference on Electronic Commerce (EC-99)},
  year      = {1999}
}

@inproceedings{Ber06,
  author       = {Daniel J. Bernstein},
  title        = {Curve25519: New Diffie-Hellman Speed Records},
  booktitle    = {Public Key Cryptography - {PKC} 2006, 9th International Conference
                  on Theory and Practice of Public-Key Cryptography},
  year         = {2006},
  url          = {https://doi.org/10.1007/11745853\_14}
}

@article{BKMS20,
  author    = {Prasad Buddhavarapu and
               Andrew Knox and
               Payman Mohassel and
               Shubho Sengupta and
               Erik Taubeneck and
               Vlad Vlaskin},
  title     = {Private Matching for Compute},
  journal   = {{IACR} Cryptol. ePrint Arch.},
  year      = {2020},
  url       = {https://eprint.iacr.org/2020/599}
}

@inproceedings{IKNPRSSSY19,
  author    = {Mihaela Ion and
               Ben Kreuter and
               Ahmet Erhan Nergiz and
               Sarvar Patel and
               Shobhit Saxena and
               Karn Seth and
               Mariana Raykova and
               David Shanahan and
               Moti Yung},
  title     = {On Deploying Secure Computing: Private Intersection-Sum-with-Cardinality},
  booktitle = {EuroS{\&}P},
  year      = {2020}
}

@inproceedings{PSWW18,
	author    = {Benny Pinkas and
	Thomas Schneider and
	Christian Weinert and
	Udi Wieder},
	title     = {Efficient Circuit-Based {PSI} via Cuckoo Hashing},
	booktitle = {EUROCRYPT},
	year      = {2018}
}

@misc{meta-ppre,
    author = {Rachad Alao and Miranda Bogen and Jingang Miao and Ilya Mironov and Jonathan Tannen},
    title = {How {Meta} is working to assess fairness in relation to race in the U.S. across its products and systems},
    howpublished={\url{https://ai.meta.com/research/publications/how-meta-is-working-to-assess-fairness-in-relation-to-race-in-the-us-across-its-products-and-systems/}}
}

@misc{pai-demog2021, 
    author={Villeneuve, Sarah and Andrus, McKane}, 
    title={{Fairer Algorithmic Decision-Making and Its Consequences: Interrogating the Risks and Benefits of Demographic Data Collection, Use, and Non-Use}}, 
    howpublished={\url{https://partnershiponai.org/paper/fairer-algorithmic-decision-making-and-its-consequences/}}, 
    journal={Partnership on AI},  
    year={2021}, month={Dec} 
}
